\documentclass{amia}
\usepackage{lipsum} %

\usepackage{etoolbox}
\patchcmd{\bibsection}{\section*{\refname}}{\section*{\centering\refname}}{}{}
\usepackage{caption}
\usepackage{subcaption}
\usepackage[T1]{fontenc}
\usepackage{bbding}
\usepackage{amsmath,amssymb,amsfonts}
\usepackage{algorithmic}
\usepackage{graphicx}
\usepackage{textcomp}
\usepackage{xcolor}
\usepackage{amsmath}
\usepackage{pifont}

\usepackage[utf8]{inputenc} %
\usepackage[T1]{fontenc}    %
\usepackage{hyperref}       %
\usepackage{url}            %
\usepackage{booktabs}       %
\usepackage{amsfonts}       %
\usepackage{nicefrac}       %
\usepackage{microtype}      %
\usepackage{algorithm}
\usepackage{bm}
\usepackage{multirow}

\usepackage{wrapfig}
\usepackage{lipsum} %

\begin{document}
\title{UQ at \#SMM4H 2023: ALEX for Public Health Analysis with Social Media}

\author{Yan Jiang, Ruihong Qiu, Yi Zhang, Zi Huang}

\institutes{
The University of Queensland\\
\{yan.jiang, r.qiu, y.zhang4, helen.huang\}@uq.edu.au
}

\maketitle
\section*{Abstract}

\textit{As social media becomes increasingly popular, more and more activities related to public health emerge. Current techniques for public health analysis involve popular models such as BERT and large language models (LLMs). However, the costs of training in-domain LLMs for public health are especially expensive. Furthermore, such kinds of in-domain datasets from social media are generally imbalanced. To tackle these challenges, the data imbalance issue can be overcome by data augmentation and balanced training. Moreover, the ability of the LLMs can be effectively utilized by prompting the model properly. In this paper, a novel ALEX framework is proposed to improve the performance of public health analysis on social media by adopting an LLMs explanation mechanism. Results show that our ALEX model got the best performance among all submissions in both Task 2 and Task 4 with a high score in Task 1 in Social Media Mining for Health 2023 (SMM4H)\cite{smm}. Our code has been released at \url{https://github.com/YanJiangJerry/ALEX}}.

\section*{Introduction}
Social media mining with natural language processing (NLP) techniques has great potential in analyzing public health information. Current techniques for public health analysis generally involve various transformer-based models, such as BERT~\cite{bert} and LLMs. With a strong ability for text processing, such models can effectively extract valuable information from social media for various kinds of tasks such as text classification.

Although existing techniques have shown remarkable performance on some of the tasks, they often encounter problems when classifying public health information on social media. Firstly, the high computational requirements for adopting an LLM in specific public health areas will increase their training difficulty. Secondly, imbalanced data also poses a significant challenge for current techniques, which may lead to biased predictions.

To address these problems, an effective method for public health text classification is urgently needed. In this paper, firstly, a complete data augmentation pipeline with weighted loss fine-tuning is adopted to address the imbalance problem. Secondly, to reduce the LLMs training cost, an explanation and correction mechanism based on LLMs is introduced. This framework is named by \textbf{A}ugmentation and \textbf{L}arge language model methods for \textbf{EX}plainable (ALEX) social media analysis about public health. The ALEX method achieves \textbf{the highest} F1 score in both \textbf{Task 2} (77.84\%) and \textbf{Task 4} (87.15\%) with a high ranking in \textbf{Task 1} (94.3\%) for test set in SMM4H 2023. Our main contributions are:

$\cdot$ A balanced training pipeline is proposed by augmentation and weighted-loss fine-tuning for imbalance problems.

$\cdot$ An LLMs explanation method is proposed to conduct post-evaluation on the predicted results without training.

\section*{Preliminaries: Task Definition}
Our UQ team has participated in three text classification tasks which are: Task 1: Binary classification of English tweets self-reporting a COVID-19 diagnosis; Task 2: Multi-class classification of sentiment associated with therapies in English tweets; Task 4: Binary classification of English Reddit posts self-reporting a social anxiety(SA) disorder diagnosis. For each task, given a set of texts from social media posts $\{x_1,...,x_n\}\subseteq \mathcal{X}$ and their corresponding labels $\{y_1,...,y_n\}\subseteq \mathcal{Y}$, the objective is to develop a model that can effectively classify the input social media texts into several predefined groups. For the following equations, bold lowercase letters denote vectors, lowercase letters denote scalars or strings and uppercase letters denote all the other functions. 

\section*{Method}
\begin{figure}[!t]
  \centering
  \includegraphics[width=0.6\textwidth]{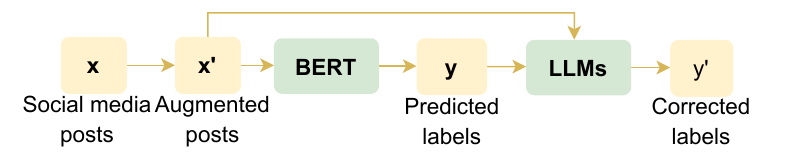}
  \caption{Overall framework of ALEX method. Firstly, the social media posts are augmented to fine-tune the BERT models. The labels are then predicted by BERT. After combining the labels with their original posts and constructing a prompt, LLMs will take the prompt to explain and correct those incorrectly predicted labels from BERT.}
  \label{fig1}
\end{figure}

In this section, the ALEX method is divided into several components as Figure \ref{fig1} shows. Firstly, balanced training involves augmentation is employed to solve the imbalance problem. Secondly, fine-tuning with the weighted loss is conducted on BERT. Lastly, an LLM is applied to explain and correct the labels predicted from BERT.

\paragraph*{Augmentation and Encoding.}
To solve the imbalanced problem, the TextAttack~\cite{SMM4H} is chosen to enrich the number of the minority class. After data augmentation, oversampling and undersampling are also introduced to make the number of each class exactly the same. After getting the balanced dataset, those texts will be encoded by BERT to get better embeddings for text classification as Equation \ref{eq2} shows. The BERT's output sequence can be divided into segments with the first token being [CLS], which is the special classification token for predicting the labels' probability.
\begin{equation} \label{eq2}
\textbf{e} = \text{BERT}(\textbf{x}'),
\end{equation}
where $\textbf{e}\in\mathbb{R}^{d}$ represent the embedding of the [CLS] token while $d$ represents its dimension. To employ the BERT's embedding for text classification, a straightforward approach is adding an additional classifier as Equation \ref{eq3} shows:
\begin{equation} \label{eq3}
\hat{y} = \text{argmax}(\text{Classifier}(\textbf{e})),
\end{equation}
where $\hat{y}\in\mathbb{R}$ is the label predicted by BERT. The output from the classifier represents the probability of each label, then the argmax function will extract the index that has the max probability, which will be the predicted label.

\paragraph*{Weighted Loss.} For Task 1 and Task 4, the F1 score for Label 1 is particularly important than Label 0. In order to ensure the model assigns more emphasis to those specific labels during the training process, the loss weight $\lambda$ will be introduced, which can be integrated into the binary cross-entropy loss as Equation \ref{eq5} shows:
\begin{equation} \label{eq5}
\ell = -\frac{1}{n} \sum_{i=1}^{n} \lambda  y  \text{log}(\hat{y}) + (1 - y) \text{log}(1-\hat{y}).
\end{equation}

\begin{wrapfigure}{r}{0.35\textwidth}
  \centering \vspace{-10pt}
  \includegraphics[width=0.35\textwidth]{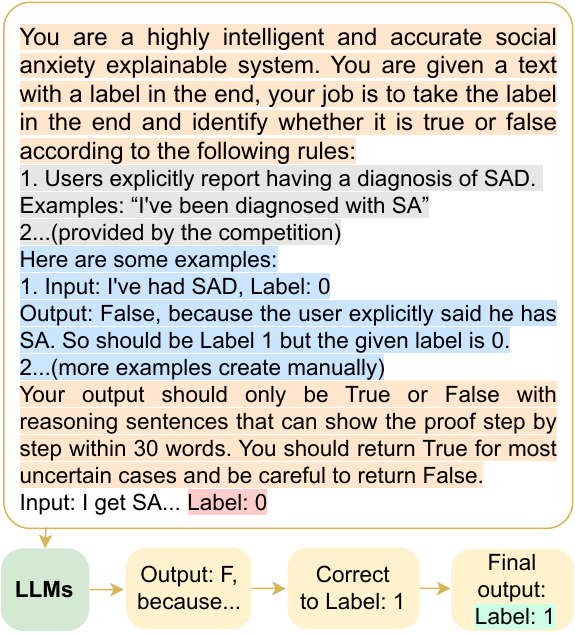}
  \caption{LLMs method. The prompt includes instructions (orange), labeling rules (grey), examples (blue), and wrong labels (red). LLMs identified the label as False thus it will be corrected (green).}
  \label{ALEX}\vspace{-30pt}
\end{wrapfigure}

\paragraph*{LLM Explanation and Corrections at Inference Time.}\label{llm}
To overcome the limited input size of BERT and enhance its performance, LLMs will be introduced. Firstly, the predicted label from BERT will be concatenated with the original text. Secondly, the processed text will be combined with the labeling rules and instructions for each task with a few representative examples to form a complete prompt for LLMs' few-shot prompting as Figure \ref{ALEX} shows:

If LLMs successfully find supporting evidence in the original text that aligns with the given label, it will return ``True'' with a concise provable explanation. However, if LLMs find the opposite proofs or fail to identify any compelling evidence that could support the given label, they will return ``False'' and will use the Chain of Thought (CoT)~\cite{chain} idea to give a step-by-step explanation as Equation \ref{eq6} shows:
\begin{equation} \label{eq6}
\text{explanation} = \text{LLMs}(\text{Prompt}(x,\hat{y})),
\end{equation}

where the explanation involves the judgments of whether LLMs think the label is true with the explanation as the proof. Prompt represents the prompt construction including the concatenation of the original text with the BERT predicted label, adding manually created labeling rules with a few examples.

For the LLMs' explanation part, a correction mechanism is designed. For cases when LLMs return ``True'', the original labels from BERT will be kept. For cases when the LLM returns ``False'', there are two options. Firstly, if the number of ``False'' cases is relatively small, the manual review can be achievable to verify the LLMs' explanation. If the evidence is compelling, the label will be corrected to the others. The second option is directly modifying the ``False''  label to the other label which will minimize the manual efforts. For binary text classification tasks, the ``False'' label can be directly converted to the other label. For the three-class text classification task, the ``False'' label can be converted to the label of the majority class.

\section*{Experiments}

\paragraph{Metrics.}
Popular evaluation metrics such as precision, recall, and F1 score are adopted. For Task 1 and Task 4, only F1 scores for Label 1 are evaluated. For Task 2, micro-averaged F1 scores (same as accuracy) are evaluated.

\vspace{5pt}
\paragraph{Implementation.}
The implementation is based on the Transformers library\footnote{\url{https://pytorch.org/docs/stable/generated/torch.nn.Transformer.html}}. For hyperparameters optimization, the batch size of \{4, 8, 16\} is tested and the AdamW optimizer is implemented with a learning rate choosing from 2e-5 to 1e-4. To mitigate overfitting problems, the weight decay from \{0.001, 0.005, 0.01\} is used. The number of warm-up steps is set to zero and the number of epochs is tested in a range of \{2, 4, 6, 8, 10, 12, 16, 20\}. For the baseline models, the batch size 8, learning rate 2e-5, weighted loss 0.1, and epoch 6 are used except for the XLNet model which uses batch size 4 due to the hardware limitations. For XLNet which has a larger token length, the inputs will not be truncated.

In Task 4, BERTweet-Large indicates that a learning rate of 2e-5, a batch size of 16, a weight decay of 0.005, and epochs of 6 will lead to the highest F1 scores for Label 1. For other tasks, weight decay of 0.01 and epoch 3 will be adopted as over-training may hinder the models' performance. For the LLMs part, GPT-3.5 is employed from OpenAI to explain and correct the BERT's prediction. For Task 1 and Task 4, the ``False'' Label 1 sample will be automatically converted to Label 0. For multi-label classifications such as Task 2, the ``False'' Label 0 predictions will be converted to the majority class which is Label 1.

\vspace{20pt}
\paragraph{Baselines.} 
The following baselines are chosen for comparison: BERT~\cite{bert}, RoBERTa~\cite{liu2019roberta}, XLNet~\cite{DBLP:journals/corr/abs-1906-08237}, BERTweet~\cite{nguyen2020bertweet} and CT-BERT (v2)~\cite{müller2020covidtwitterbert}.
In overall comparison, the ``-B'' indicates the base models, and ``-L'' denotes the large models. Only BERTweet-Large is used for ALEX-Large in Task 2 and Task 4 while CT-BERT (v2) is only used for Task 1 as it is the default baseline for COVID-19 classification on social media dataset.

\subsection*{Overall Performance}
As shown in Table \ref{tab3}, for Task 1, the original BERT models may outperform large models such as XLNet, which may owing to the serious impact of the imbalanced datasets. Generally, for BERT-related models, the large model may outperform the base model. However, XLNet-Large underperforms XLNet-Base and gets a zero score for Label 1's F1 score in Task 1 which may related to the complexity of the model. For RoBERTa, experiment results show that it can get better results compared to BERT on Task 1 and Task 4. However, for the base model, RoBERTa underperforms BERT in Task 2. It is noticeable to see that BERTweet-Large model performances are generally better than others on the F1 score for Label 1 in Task 2 and Task 4. Moreover, for Task 1, CT-BERT (v2) gets the highest F1 score for Label 1 compared to all other baseline methods as it has already been further fine-tuned on COVID-19-related Twitter datasets.

\begin{wrapfigure}{r}{6.2cm}\vspace{-170pt}
 \begin{tabular}{cc}
    \begin{subfigure}[b]{0.17\textwidth}
      \includegraphics[width=\textwidth]{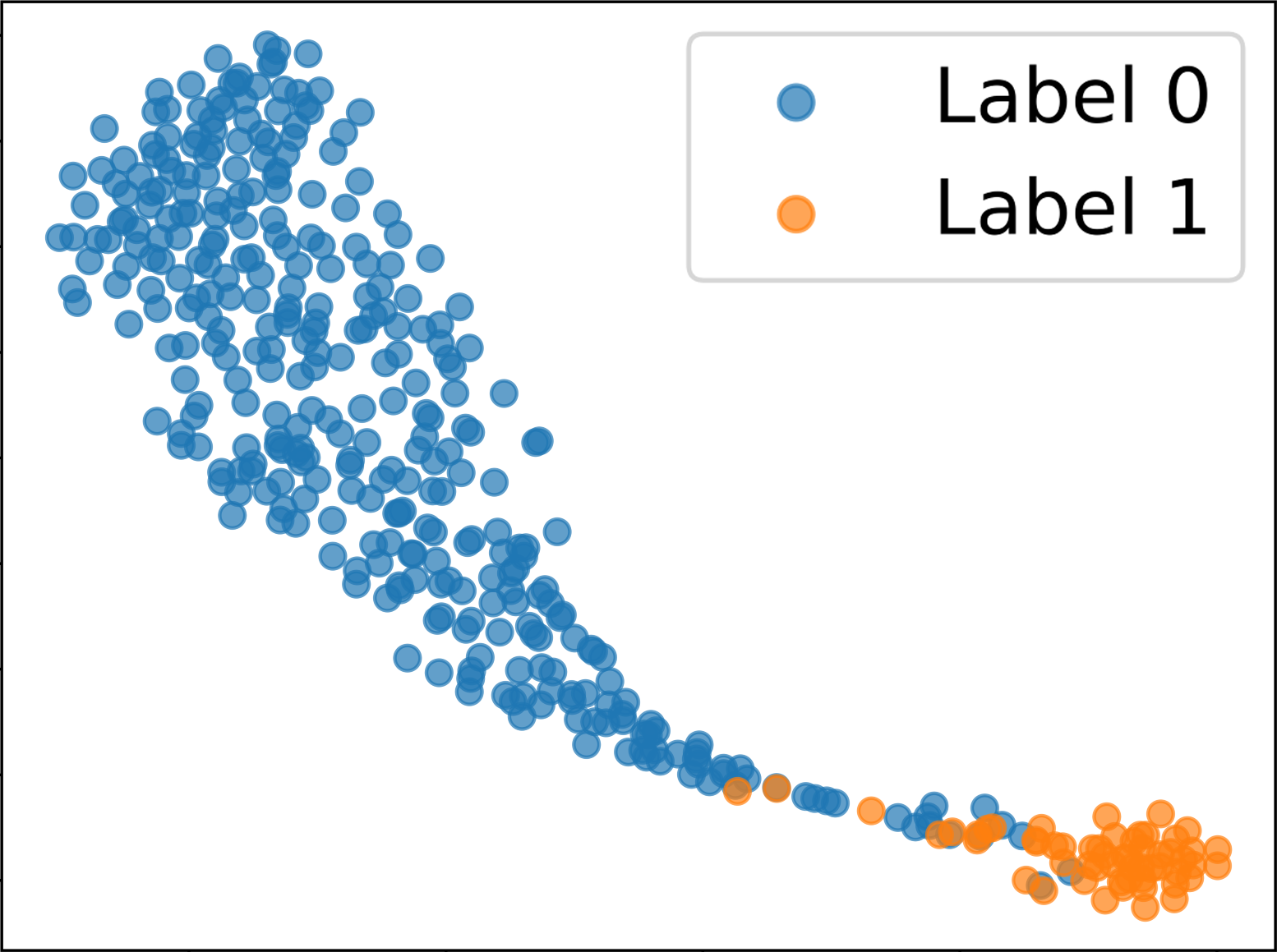}
      \caption{Task 2.}
      \vspace{5pt}
    \end{subfigure}
    &
    \begin{subfigure}[b]{0.17\textwidth}
      \includegraphics[width=\textwidth]{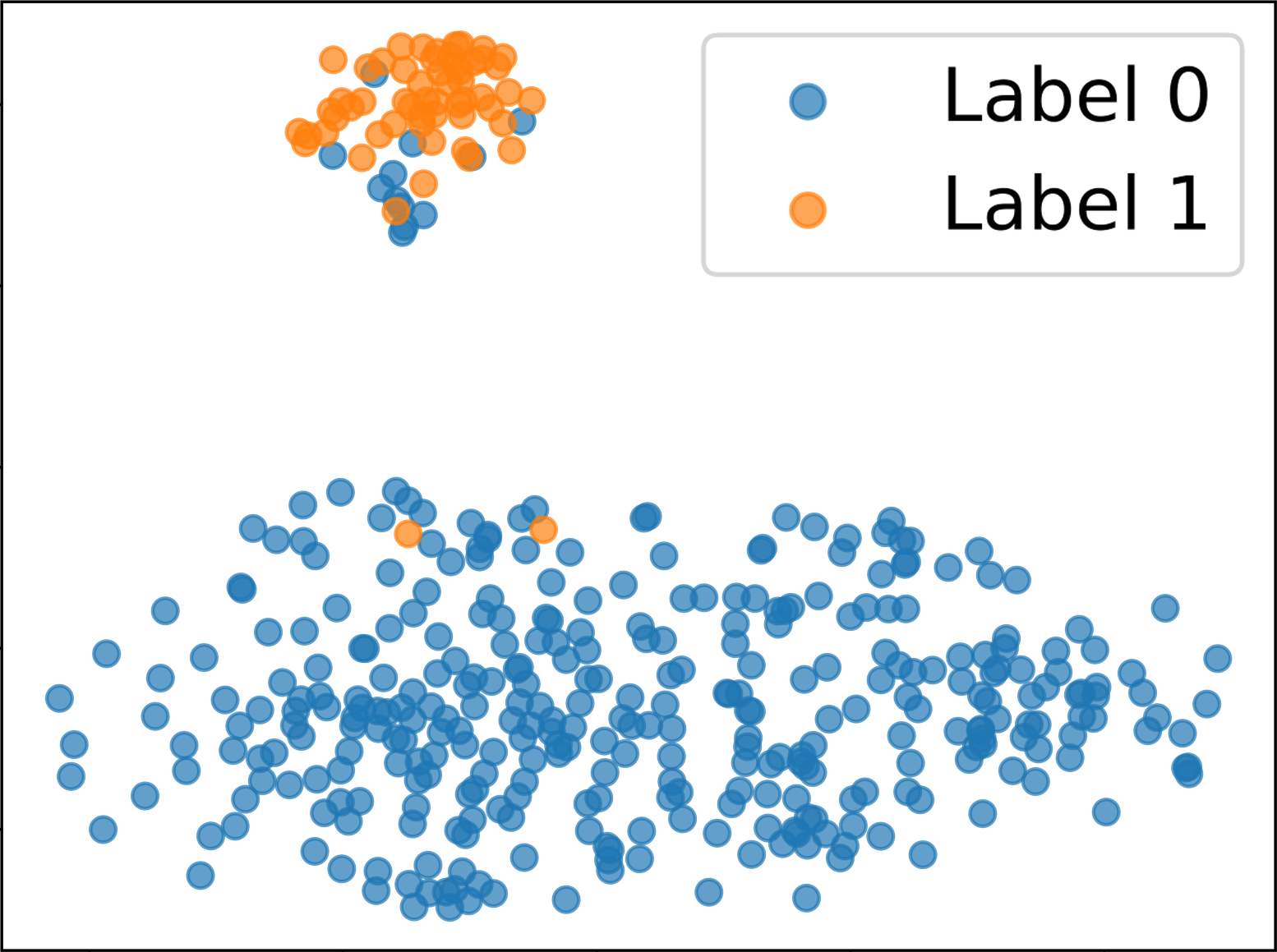}
      \caption{Task 2 (Balanced).}
      \vspace{5pt}
    \end{subfigure}
  \end{tabular}
  
  \begin{tabular}{cc}
    \begin{subfigure}[b]{0.17\textwidth}
      \includegraphics[width=\textwidth]{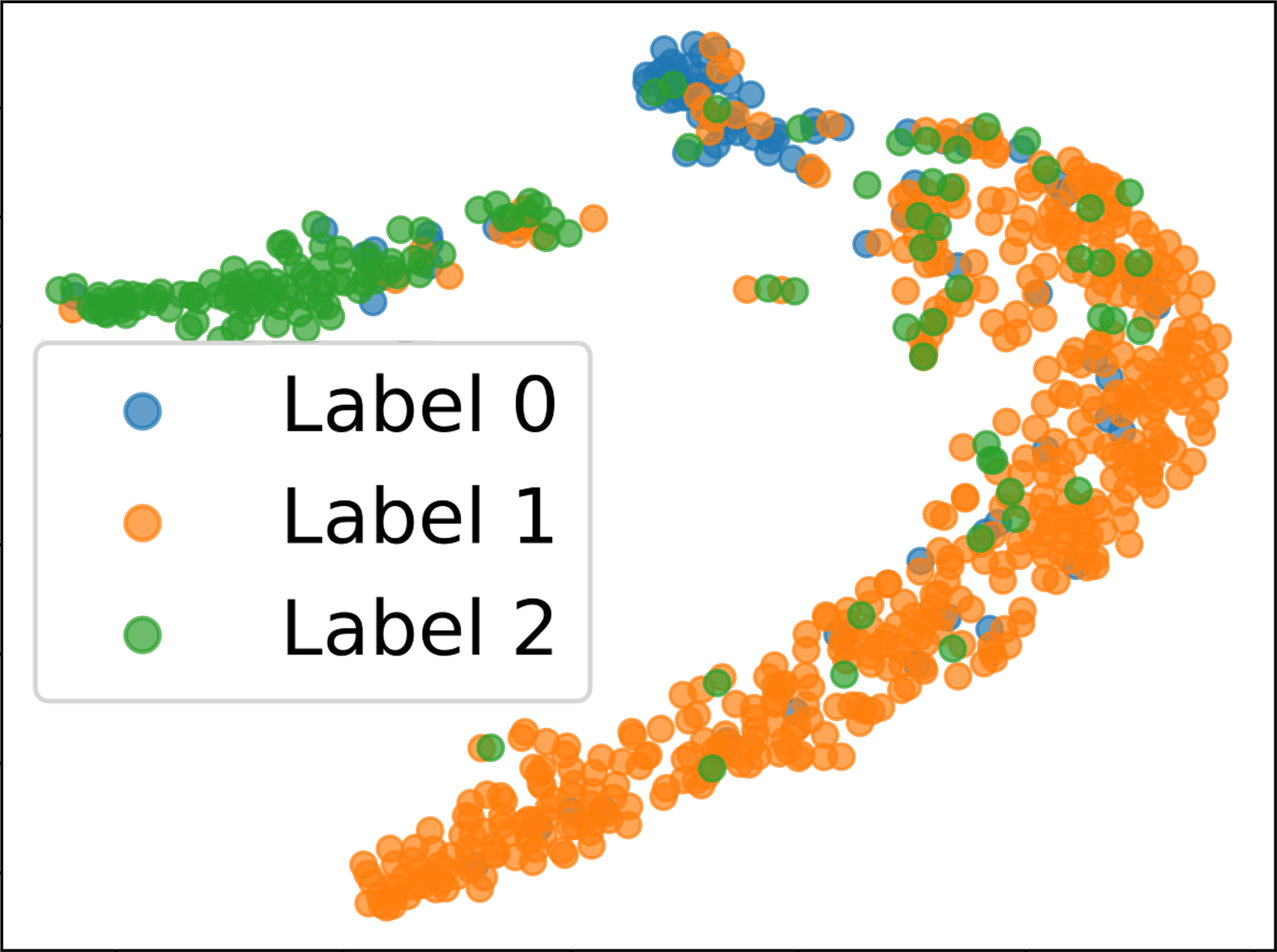}
      \caption{Task 2.}
      \vspace{5pt}
    \end{subfigure}
    &
    \begin{subfigure}[b]{0.17\textwidth}
      \includegraphics[width=\textwidth]{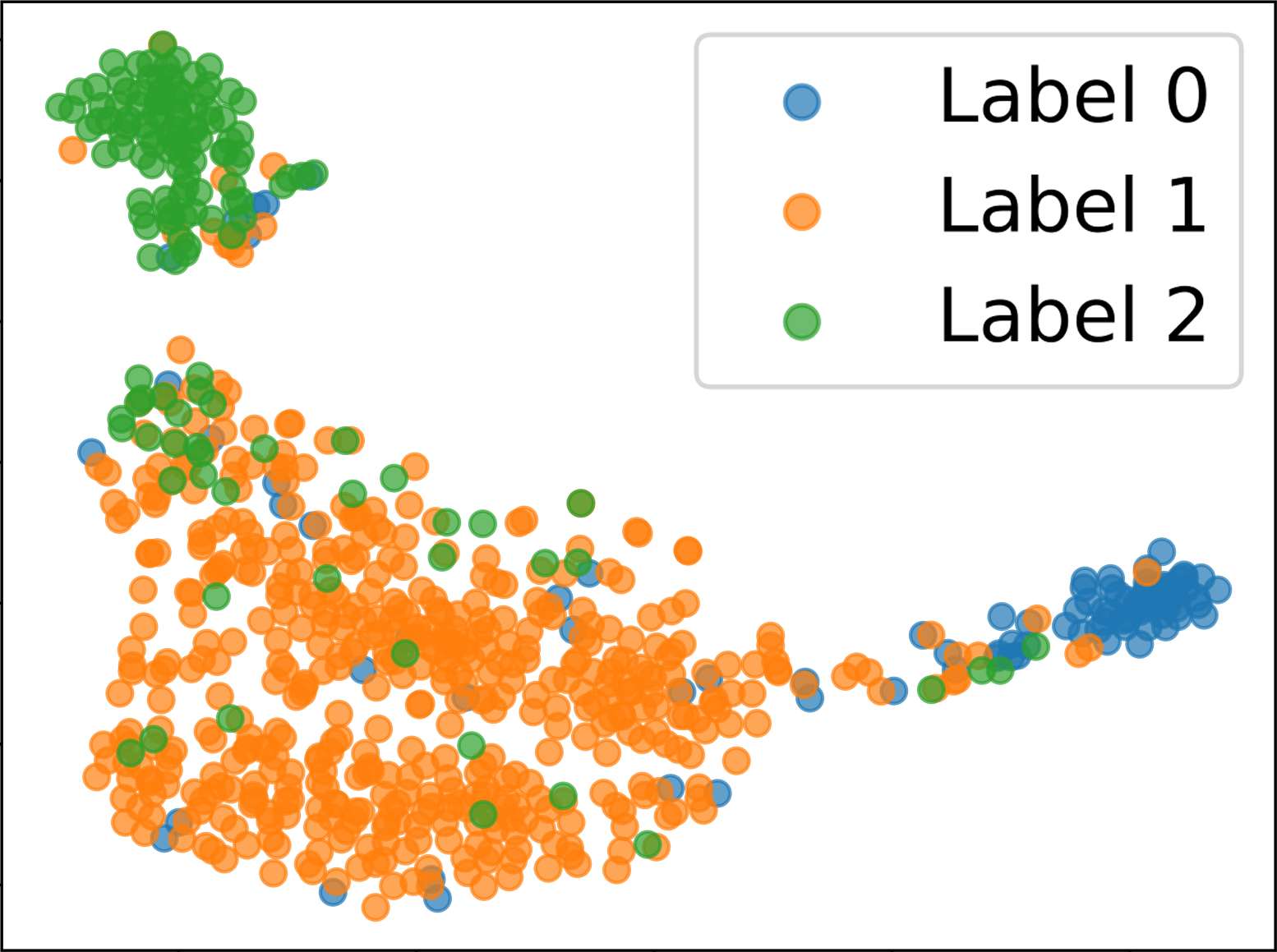}
      \caption{Task 2 (Balanced).}
      \vspace{5pt}
    \end{subfigure}
  \end{tabular}
  
  \begin{tabular}{cc}
    \begin{subfigure}[b]{0.17\textwidth}
      \includegraphics[width=\textwidth]{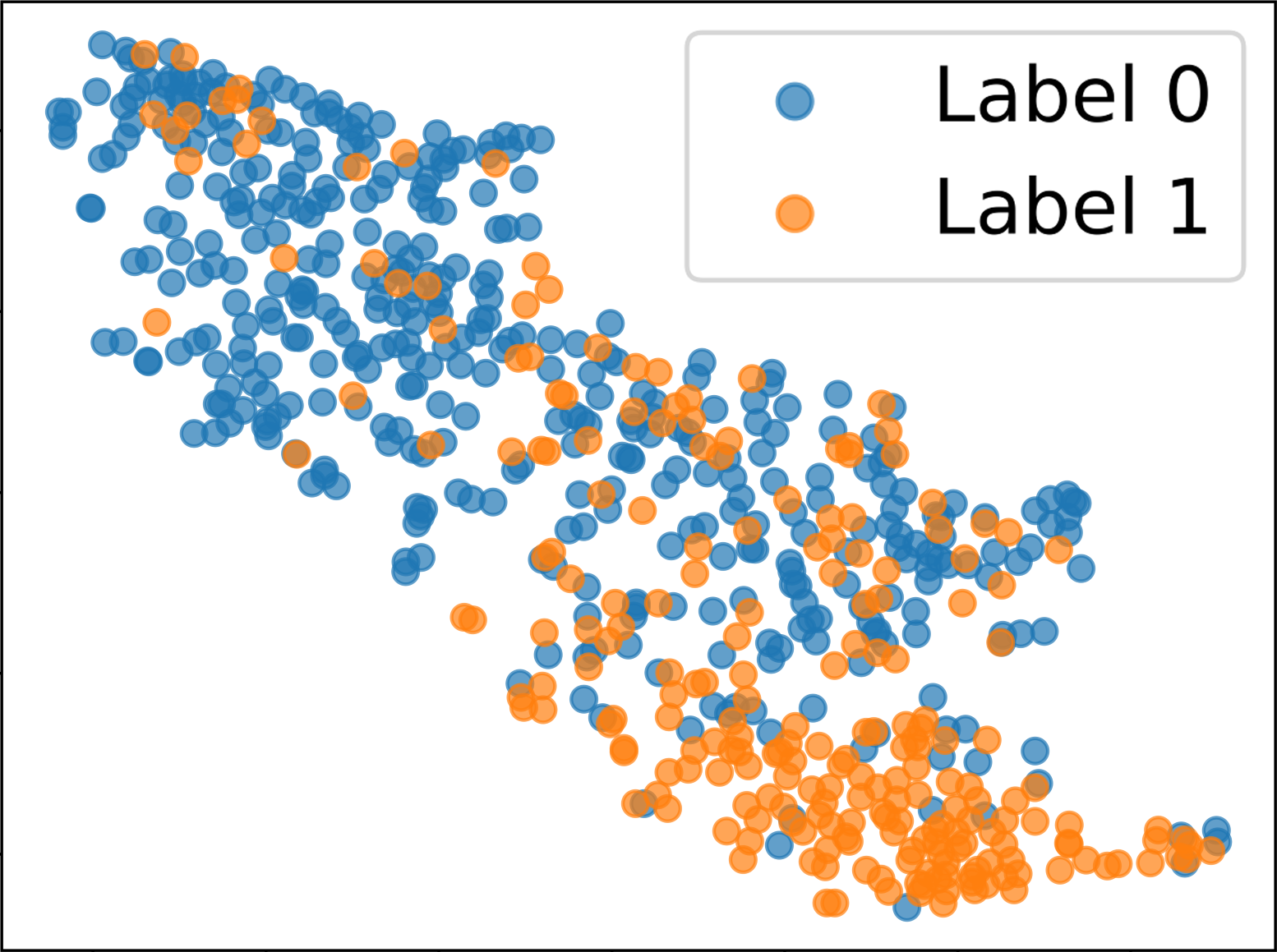}
      \caption{Task 4.}
    \end{subfigure}
    &
    \begin{subfigure}[b]{0.17\textwidth}
      \includegraphics[width=\textwidth]{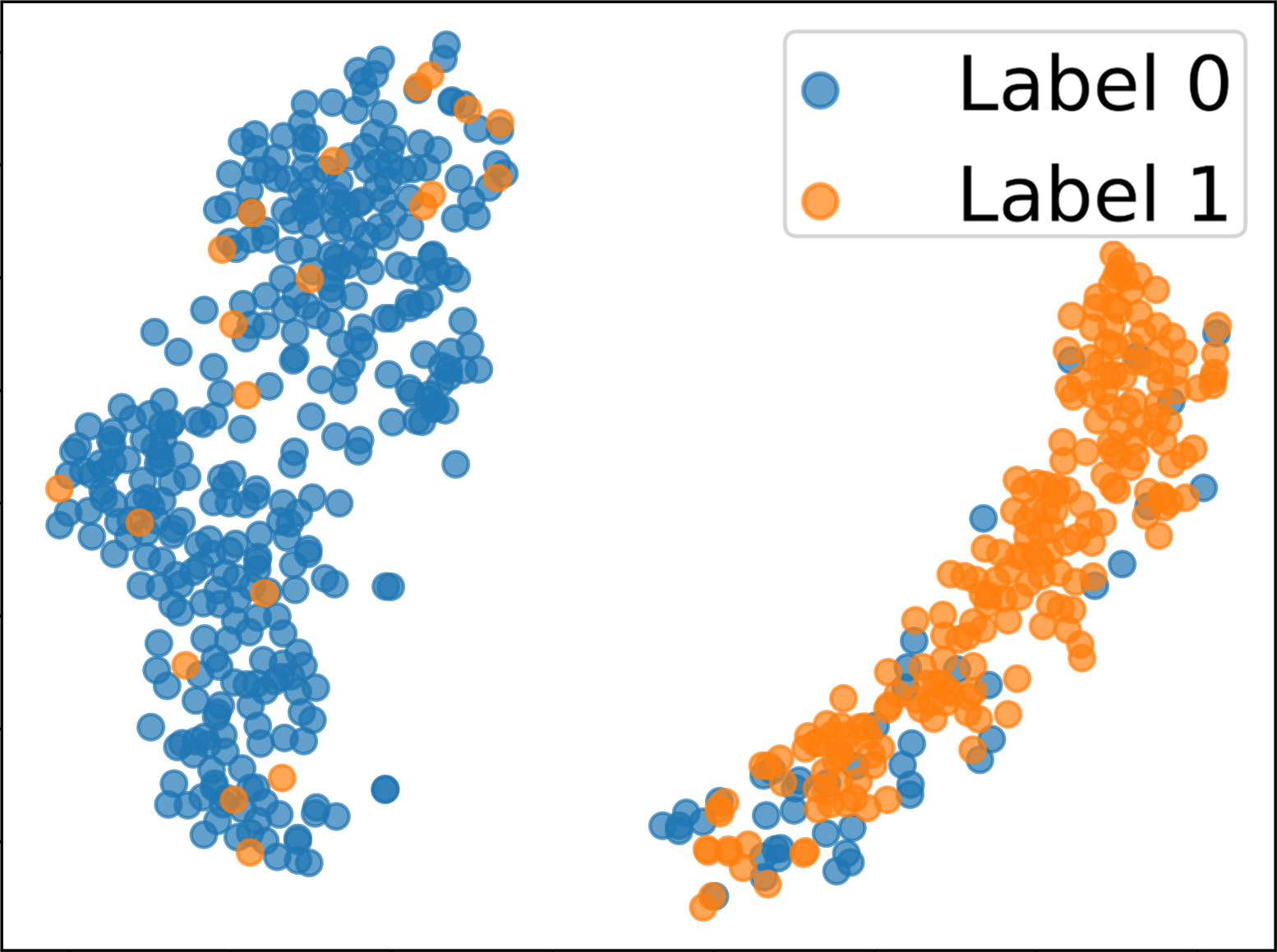}
      \caption{Task 4 (Balanced).}
    \end{subfigure}
  \end{tabular}
  \caption{Visualisations of Embeddings.}
  \label{tsne}
  \vspace{-30pt}
\end{wrapfigure}

Our ALEX-Large method outperforms all other methods as it integrates balanced training and LLMs correction. For Task 2 and Task 4, the t-SNE visualizations of the [CLS] embeddings after the baselines' imbalanced training method (left-hand side) compared to our balanced training method (right-hand side) are shown in Figure \ref{tsne}. It is obvious that for most tasks such as Task 2 and Task 4, the embeddings are well separated compared to baseline methods. Therefore, our method can be proven to enlarge the inter-class distance between different labels and separate different classes better than baseline methods in social media classification tasks related to public health.

\begin{wraptable}{l}{6.5cm}
\captionsetup{justification=raggedright}
\caption{Overall model performance.}\label{tab3}
\resizebox{1.5\linewidth}{!}{
\begin{tabular}{c|ccc|ccc|ccc}
\toprule
\multirow{2}{*}{Model} & \multicolumn{3}{c|}{Task 1} & \multicolumn{3}{c|}{Task 2}& \multicolumn{3}{c}{Task 4}\\
\cline{2-10}
& F1  &  Accuracy & M-Avg  &  F1  & Accuracy & M-Avg & F1 & Accuracy & M-Avg \\
\midrule
BERT-B & 82.25 & 93.15 & 93.74 & 83.33 & 71.24 & 58.94 & 79.75 & 84.15 & 83.37\\
\midrule
BERT-L & 85.75 & 95.02 & 93.79 & 80.44 & 69.30 & 64.92 & 81.17 & 85.74 & 84.85\\
\midrule
RoBERTa-B & 91.84 & 92.77 & 91.01 & 77.33 & 70.07 & 66.45 & 81.86 & 87.13 & 85.87\\
\midrule
RoBERTa-L & 92.71 & 94.97 & 93.56 & 81.35 & 74.45 & 71.07 & 44.52 & 72.74 & 63.22\\
\midrule
XLNet-B & 3.04 & 64.56 & 51.47 & 84.82 & 76.28 & 76.87 & 78.92 & 82.73 & 82.14 \\
\midrule
XLNet-L & 0.0 & 64.33 & 50.37 & 85.00 & 75.74 & 76.78 & 84.89 & \underline{89.18} & \underline{87.66}\\
\midrule
BERTweet-B & 88.31 & 93.67 & 92.82 & 79.97 & 72.44 & 68.37 & 80.77 & 83.49 & 82.93\\
\midrule
BERTweet-L & 91.22 & 95.57 & 93.61 & \underline{85.61} & 73.67 & 77.49 & \underline{85.42} & 87.01 & 85.38\\
\midrule
CT-BERT (v2) & \underline{93.11} & \underline{95.62} & \underline{94.47} & 85.02 & \underline{76.43} & \underline{77.81} & 61.93 & 70.29 & 68.78 \\
\midrule
\midrule 
ALEX-B (ours) & 90.77 & 92.04 & 91.25 & 82.20 & 73.00 & 71.19 & 83.39 & 88.25 & 87.15\\
\midrule
ALEX-L (ours) & \textbf{94.97} & \textbf{96.71} & \textbf{95.84} & \textbf{89.13} & \textbf{77.84} & \textbf{79.57} & \textbf{88.17} & \textbf{89.84} & \textbf{88.26}\\
\bottomrule
\end{tabular}
}
\end{wraptable}

\subsection*{Ablation Study}
In order to validate the effectiveness of ALEX's method, an ablation study is conducted. As shown in Table \ref{tab6}, firstly, the effectiveness of balanced training (``Bal'') is verified by comparing balanced training to baselines that use imbalanced training. 

\begin{wraptable}{r}{6cm}
\centering \vspace{12pt}
\caption{Ablation study for ALEX.}\label{tab6}
\resizebox{\linewidth}{!}{
\begin{tabular}{c|c|cc|cc|cc}
\toprule
\multirow{2}{*}{Bal} & \multirow{2}{*}{LLMs} & \multicolumn{2}{c|}{Task 1} & \multicolumn{2}{c|}{Task 2}& \multicolumn{2}{c}{Task 4}\\
\cline{3-8}
& & F1 & Acc & F1 & Acc & F1 & Acc \\
\midrule
\XSolid & \XSolid & 93.11 & 95.62 & 85.61 & 73.67 & 85.42 & 87.01\\
\midrule
\Checkmark & \XSolid & \textbf{94.97} & \textbf{96.71} & 86.17 & 74.34 & 86.64 & 87.94\\
\midrule
\Checkmark & \Checkmark & 93.24 & 95.11 & \textbf{89.13} & \textbf{77.84} & \textbf{88.17} & \textbf{89.84}\\
\bottomrule
\end{tabular}
}\vspace{-8pt}
\end{wraptable}
Another component is the LLMs explanation and correction (``LLMs''). We constrain the GPT-3.5 to only explain and correct Label 1's samples from BERT's prediction to improve Label 1's F1 score. For Task 2, GPT-3.5 is constrained to only verify Label 0 predictions and correct to the majority class Label 1. Results reveal that GPT-3.5's correction can improve Task 2 and Task 4 Label 1's F1 scores from 86 to above 88 compared to the results after balanced training. For Task 1, only the balanced training from ALEX is used.

\section*{Conclusions}
In this paper, an ALEX framework is proposed to improve model performance for public health datasets on social media. The remarkable competition results from SMM4H 2023 have proven the effectiveness of our ALEX method.

\paragraph*{Acknowledgements}
This work is supported by Developing a proof-of-concept self-contact tracing app to support epidemiological investigations and outbreak response (Australia-Korea Joint Research Projects - ATSE Tech Bridge Grant).

\bibliographystyle{vancouver}
\bibliography{amia}  
\end{document}